# Efficient Intrinsically Motivated Robotic Grasping with Learning-Adaptive Imagination in Latent Space

Muhammad Burhan Hafez, Cornelius Weber, Matthias Kerzel and Stefan Wermter
*Knowledge Technology, Department of Informatics, University of Hamburg, Germany*
{hafez, weber, kerzel, wermter}@informatik.uni-hamburg.de

*Abstract*—Combining model-based and model-free deep reinforcement learning has shown great promise for improving sample efficiency on complex control tasks while still retaining high performance. Incorporating imagination is a recent effort in this direction inspired by human mental simulation of motor behavior. We propose a learning-adaptive imagination approach which, unlike previous approaches, takes into account the reliability of the learned dynamics model used for imagining the future. Our approach learns an ensemble of disjoint local dynamics models in latent space and derives an intrinsic reward based on learning progress, motivating the controller to take actions leading to data that improves the models. The learned models are used to generate imagined experiences, augmenting the training set of real experiences. We evaluate our approach on learning vision-based robotic grasping and show that it significantly improves sample efficiency and achieves near-optimal performance in a sparse reward environment.

## I. INTRODUCTION

Autonomous learning of robotic motor behavior requires finding a mapping from raw sensory input to raw motor output that defines a desired behavior without any prior knowledge of the task. Deep Reinforcement Learning (RL) enables learning such behaviors from trial-and-error experience using deep neural networks as function approximators. It has recently shown great success in learning complex control behaviors, achieving super-human performance in playing Atari 2600 games [1] and acquiring a range of robotic manipulation skills [2]. This success has come at the expense of high sample complexity that limits how fast the robot can learn good control policies, which is particularly problematic in the presence of high-dimensional noisy observations and real-time constraints often found in realistic robotic domains.

To address this problem, different approaches have been proposed, such as sampling experiences with probability proportional to their temporal difference error instead of random sampling [3], counting unsuccessful episodes of policy execution as successful ones by replaying each learning episode with different goals than the one the robot was attempting to achieve in a multi-goal RL setting [4], [5], or extending the classic count-based exploration methods from tabular to deep neural network representations [6], [7].

In contrast, works on intrinsic motivation have addressed the problem mostly by looking at how the robot's evolving knowledge of its world dynamics can be utilized by self-rewarding action choices leading to novel observations where high extrinsic rewards might be gained. Since this actively directs the exploration using a learned dynamics model from highly to less predictable regions of the state space, it is especially useful for learning in sparse reward environments that lack the important feedback on how to improve exploration.

Examples of intrinsic reward measures include model prediction error [8], [9], learning progress [10]–[12], competence progress at achieving self-generated goals [13], [14], and change in policy value [15], [16]. While most of these approaches exploit the useful information a learned dynamics model offers, they almost only employ it for computing the intrinsic reward without leveraging its potential for model-based learning of policy and value functions.

Predictive world models are typically known for their ability to boost the sample efficiency of RL methods. In particular, they allow for imagining experiences by making predictions about future states and rewards which can then be used for policy learning, reducing the number of required real experiences of costly agent-environment interactions. Moreover, using features extracted from a recurrent predictive world model as input to an RL agent has been found to achieve state-of-the-art results on challenging RL tasks [17]. This further confirms the significance of imagination, since the extracted features contain information about the future.

Imagination, defined as mental simulation of motor behavior, is considered strong evidence for cognitive synergy as it requires a combination of abstract perceptual and motor representations, episodic and working memory and mental manipulation of representations [18]. This imagination-centered synergy has been further distinguished neurally by examining the different brain regions activated during imagination, involving cognitive and motor areas [19], [20] and is a clear example of cognitive development in children where increasingly complex behaviors develop from the recombination of existing, less complex behaviors. Imagination is also essential to mental practice which is the cognitive rehearsal of a physical skill and found to facilitate skill acquisition [21].

It has been argued that automating imagination holds the potential for advancing deep learning beyond finding correlations in data and for expanding the focus from problem-solving to problem creation, since imagination involves the ability to be exploratory, novelty seeking, and curious [22]. More recently, deep RL methods that employ imagination have been shown to share a number of similarities with human mental simulation, which plays a key role in human cognition, particularly the capacity to build mental models from remembered experiences and using them in decision-making [23].

The approaches for incorporating imagination into deep RL can be grouped into two categories: (i) augmenting the training set of real experiences with imagined experiences [24], [25]; (ii) integrating online planning with model-free value estimation [26]–[29]. A major issue in these approaches is that they assume a perfect world model. In complex domains, model prediction errors are inevitable and can quickly compound during action planning leading to useless long-term predictions. This is one of the reasons model-based RL algorithms failed to reach the performance of their model-free counterparts in such domains.

In our previous work [12], we proposed an algorithm called Deep Intrinsically motivated Continuous Actor-Critic (Deep ICAC) that incrementally learns an ensemble of disjoint local dynamics models in a learned latent space and derives an intrinsic reward based on learning progress. In this paper, we propose a learning-adaptive imagination approach that generates training sequences of imagined experiences. These sequences are generated only in local sensory regions where the corresponding local dynamics models retain high prediction accuracy. This allows the imagination to continually adapt to changes in learning dynamics models. We integrate our approach with the Deep ICAC algorithm. This integration makes the imagination depth itself adaptive, since it can vary according to the accuracy of the local models of the regions it spreads over. We evaluate our approach on learning robotic grasping from raw pixels and sparse rewards and show that it improves sample efficiency and achieves near-optimal performance.

## II. RELATED WORK

Combining model-free and model-based RL is a well-studied problem. One of the earliest works in this direction is Dyna-Q [30] which learns an action-value function from both real and model-generated experiences. Over the last five years, there has been a growing interest in developing Dyna-like methods in deep RL. For example, Gu et al. [24] augment a replay buffer of past state transitions with imagined transitions generated by a learned model to speed up model-free RL. They iteratively refit a linear model to recently collected transitions and generate short imagined rollouts from states sampled from these transitions. While they attempt to reduce model bias by sampling from regions where the model has been trained, the learned linear model is not expressive enough to represent complex dynamics and generate imagined rollouts in control tasks from raw-pixel input, as the authors indicate. In contrast to [24], Kalweit and Boedecker [25] use imagined transitions for updating the value and policy functions only when there is a high uncertainty in action-value predictions, as estimated using bootstrap. The approach is shown to improve the efficiency of learning continuous control tasks, but it does not address the prediction errors of the model and requires training of additional critic networks for bootstrap uncertainty estimation.

Racanière et al. [26] follow a different path by using imagination as a context for model-free value estimation. This is done by encoding rollouts of imagined observations with a recurrent neural network. The encoded rollouts are concatenated and used as an additional input to the value and policy functions. In another work, a model-based controller is used such that it randomly generates a number of candidate action sequences at each timestep, simulates the imagined trajectories with a learned model, and executes the first action of the trajectory yielding the highest reward [27]. The controller is then used to initialize the policy of a model-free RL agent with supervised data by providing it with target actions at some sampled states, which is found to make the chosen model-free RL method more sample-efficient. The use of a model predictive controller based on random-sampling, however, limits the application of the approach to low-dimensional action spaces and short planning horizons.

Unlike previous works, Feinberg et al. [28] decompose value estimate into a part with imagined rewards predicted by a dynamics model over a short horizon and a subsequent part estimated by a model-free critic. Their method, called Model-based Value Expansion (MVE), is shown to boost the sample efficiency of learning, but on control tasks with low-dimensional observations (<20). MVE avoids issues related to learning from data generated with an outdated model by not using an imagination buffer. However, it relies on the strong assumption that the model is accurate over a short, fixed horizon, which is very likely to fail in practice, since the model can generate noisy data in the early learning while still being trained jointly with the target policy, and a measure of model inaccuracy becomes necessary.

While these works incorporate imagination irrespective of the prediction error of a learned dynamics model, our proposed approach takes this error into account before initiating and while performing imagination, by using spatially and temporally local estimates of prediction accuracy.

## III. BACKGROUND

### A. Reinforcement Learning

We consider a standard RL problem where the goal is to learn a policy $\pi : S \to P(A)$, a mapping from states $S$ to probability distributions over actions $A$, that maximizes expected return $R_t = \sum_{i=t}^{T-1} \gamma^{i-t} r(s_i, a_i)$ under $\pi$ and the world dynamics, where $r : S \times A \to \mathbb{R}$ is the reward function and $\gamma \in [0,1]$ is the discount factor. A value function is defined as follows: $V^\pi(s_t) = \mathbb{E}[R_t \mid s_t]$ and the objective becomes to find the optimal policy $\pi^*$ that satisfies:

$$\pi^* = \arg\max_\pi V^\pi(S_0) \qquad (1)$$

where $S_0 \subseteq S$ is a set of initial states.

Actor-critic algorithms in RL are well suited for continuous action spaces since they learn a policy and a value function simultaneously. Two example algorithms are *Deep Deterministic Policy Gradient* (DDPG) [31] and *Continuous Actor-Critic Learning Automaton* (CACLA) [32]. In the following, we discuss CACLA which is used as the base RL algorithm in our work.

### B. Continuous Actor-Critic Learning Automaton

CACLA is a model-free on-policy actor-critic algorithm. It approximates the policy function and the value function using two neural networks: an actor $Ac(\cdot \mid \theta^{Ac})$ and a critic $V(\cdot \mid \theta^V)$ with parameters $\theta^{Ac}$ and $\theta^V$ respectively. The critic learns an approximate value function from sampled transitions by minimizing the temporal difference (TD) loss:

$$V_{target} = r + \gamma V'(s' \mid \theta^{V'}) \qquad (2)$$

$$\mathcal{L}_{TD}(\theta^V) = \mathbb{E}_{(s,a,r,s') \sim B}\left[\left(V_{target} - V(s \mid \theta^V)\right)^2\right]$$

where $B$ is a replay buffer of previously observed state transitions $(s, a, r, s')$, and $V'(. \mid \theta^{V'})$ is the critic's target network with parameters $\theta^{V'}$ slowly moving toward their corresponding parameters of the critic's network, as follows: $\theta^{V'} \leftarrow \tau \theta^V +$

$(1 − τ)θ^{V'}$, with $τ ≪ 1$. The target network was first introduced in [1] to provide sufficiently stable targets desirable for training deep neural networks.

The probability of selecting action $a$ at state $s$ according to policy $π$ is defined depending on the chosen distribution, e.g. Gaussian centered at $Ac(s|θ^{Ac})$ with standard deviation $σ$:

$$π(a|s) = \frac{1}{\sqrt{2π}σ} e^{-(a - Ac(s|θ^{Ac}))^2 / 2σ^2} \quad (3)$$

The actor is updated toward an exploratory action $a$ only when the resulting TD-error is positive. This is because when the action $a$ results in an increase in the critic's current estimate of the value of a given state $s$ (TD-error > 0), this action is judged to be better that the actor's current estimation of the optimal action and thus the actor is updated to make its output closer to $a$. Formally, the actor is updated to minimize the loss:

$$\mathcal{L}_{actor}(θ^{Ac}) = \mathbb{E}_{(s,a,r,s') \sim B} \left[ (a - Ac(s|θ^{Ac}))^2 \mid δ(s,r,s') > 0 \right] \quad (4)$$

where $δ(s,r,s') = r + γV'(s'|θ^{V'}) - V(s|θ^V)$ is the TD-error. This policy update is different from gradient ascent on the value (the critic's output) used in other algorithms, like DDPG, for updating the policy, which is prone to divergence if the critic is not fully trained, as observed in [32].

## IV. APPROACH

In this section, we present our approach for integrating learning-adaptive imagination with the Deep ICAC algorithm [12]. We first explain the core components of Deep ICAC and show where imagination fits in the learning process, and then present how imagination is adaptively used to increase the sample efficiency of the algorithm.

### A. Local dynamics-based intrinsic motivation

**Latent representation learning:** We build upon the Deep ICAC algorithm which learns an ensemble of local dynamics models and generates an intrinsic reward based on learning progress. It uses CACLA to learn the target policy. A latent representation is learned by jointly minimizing a combined convolutional autoencoder's reconstruction and value prediction loss:

$$\mathcal{L}_{rec}(\widetilde{ω}, ω) = \left\| g\left(φ_{s_t}|\widetilde{ω}\right) - s_t \right\|_2^2$$
$$\mathcal{L}_{critic}(θ^V, ω) = \left(y_t - V(s_t|ω, θ^V)\right)^2 \quad (5)$$
$$\mathcal{L}_{combined}(ω, θ^V, \widetilde{ω}) = λ_{rec}\mathcal{L}_{rec}(\widetilde{ω}, ω) + λ_{critic}\mathcal{L}_{critic}(θ^V, ω)$$

where $φ_{s_t} = f(s_t | ω)$ is the latent state representation at timestep $t$, $f(\cdot|ω)$ and $g(\cdot|\widetilde{ω})$ are the encoder and decoder networks with parameters $ω$ and $\widetilde{ω}$ respectively, $y_t = r_t + γV'(s_{t+1}, |ω', θ^{V'})$ is the target value with $V'(\cdot|ω', θ^{V'})$ being the critic's target network parametrized by $(ω', θ^{Q'})$, and $λ_{rec}$ and $λ_{critic}$ are weighting constants on the individual loss terms. The learned latent representation is fed as input to the actor network. By sharing the learning parameters between the encoder and the critic, the jointly optimized representation is learned to be a good state discriminator and value predictor. The learning architectures of the critic and actor are shown in Fig. 1(a) and Fig. 1(b) respectively.

**Latent space self-organization:** In Deep ICAC, the space of learned latent representations is self-organized during exploration into local regions with local dynamics models with the help of an incremental Self-Organizing Map (SOM). Particularly, we use the *Instantaneous Topological Map* (ITM) [33], following [12]. This is because ITM is originally designed for strongly correlated stimuli, which is the case here since the stimuli are generated by exploration of the state space along continuous trajectories, and has been successfully used in RL [34], [12]. The ITM is defined by a set of nodes $i$, each with a weight vector $w_i$, and a set of edges connecting each node $i$ to its neighbors $N(i)$. It starts with two connected nodes, and when a new stimulus $φ_s$ is observed, the following adaptation steps are performed:

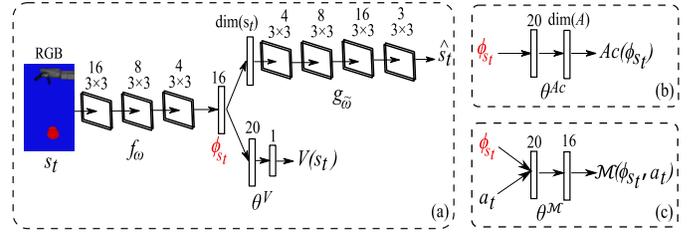

Fig. 1. Actor-critic and latent dynamics architectures: (a) A fully convolutional autoencoder that takes in a raw image $s_t$ and generates a reconstruction $\hat{s}_t$ is jointly trained with the critic and consists of 7 convolutional and 2 dense layers. The number and size of the convolutional filters used are shown above the corresponding layers; (b) The actor is a feedforward network with 2 dense layers whose output dimensionality is dim($A$), where $A$ is the action space. It takes as input the 16-D latent representation $φ_{s_t}$ trained to minimize the combined critic and reconstruction loss; (c) The latent dynamics model is a feedforward network that takes as input the current state's latent representation and the current action. It has one hidden dense layer followed by a dense output layer that outputs a prediction of the latent representation at the next timestep.

1. Matching: Find nearest node $n$ and second nearest node $n'$ to $φ_s$: $n \leftarrow \arg\min_i \|φ_s - w_i\|_2^2$, $n' \leftarrow \arg\min_{j, j \neq n} \|φ_s - w_j\|_2^2$.
2. Edge adaptation: Create an edge between $n$ and $n'$ if they are not connected. Check, for all nodes $m$ in $N(n)$, if $n'$ lies inside the Thales sphere through $m$ and $n$ (i.e. $(w_n - w_{n'}) \cdot (w_m - w_{n'}) < 0$). If true, remove the edge between $n$ and $m$, and then, if $m$ has no remaining edges, remove $m$.
3. Node adaptation: If $φ_s$ lies outside the Thales sphere through $n$ and $n'$, i.e. $(w_n - φ_s) \cdot (w_{n'} - φ_s) > 0$, and if $\|φ_s - w_n\|_2^2 > e_{max}$, where $e_{max}$ is a given threshold, create a new node $v$ with $w_v = φ_s$ and an edge with $n$.

An example of an approximate dynamics model $\mathcal{M}(\cdot, \cdot | θ^{\mathcal{M}})$, which predicts the next state encoding given the current action and state encoding and is trained to minimize the loss $\left\| \mathcal{M}(φ_{s_t}, a_t|θ^{\mathcal{M}}) - φ_{s_{t+1}} \right\|_2^2$, is shown in Fig. 1(c). In this work, in order to generate a complete imagined experience, we additionally learn a latent reward function $\mathcal{R}(\cdot, \cdot | θ^{\mathcal{R}})$ which predicts the immediate reward and is trained to minimize the loss $\left\| \mathcal{R}(φ_{s_t}, a_t|θ^{\mathcal{R}}) - r_t \right\|_2^2$. Each region $n$ of the latent space (node in ITM) is assigned a separate local dynamics model $\mathcal{M}_n$ and reward function $\mathcal{R}_n$. During learning, we maintain a moving average of the combined prediction error of $\mathcal{M}_n$ and $\mathcal{R}_n$ of the region over a window of $σ$ recent predictions:

$$\langle e_t^{prd} \rangle = \frac{1}{σ} \sum_{i=1}^{σ} e_i^{prd} \mid_{e_i^{prd} = \left\| \mathcal{M}_n(φ_{s_i}, a_i|θ^{\mathcal{M}}) - φ_{s_{i+1}} \right\|_2^2 + \left\| \mathcal{R}_n(φ_{s_i}, a_i|θ^{\mathcal{R}}) - r_i \right\|_2^2} \quad (6)$$

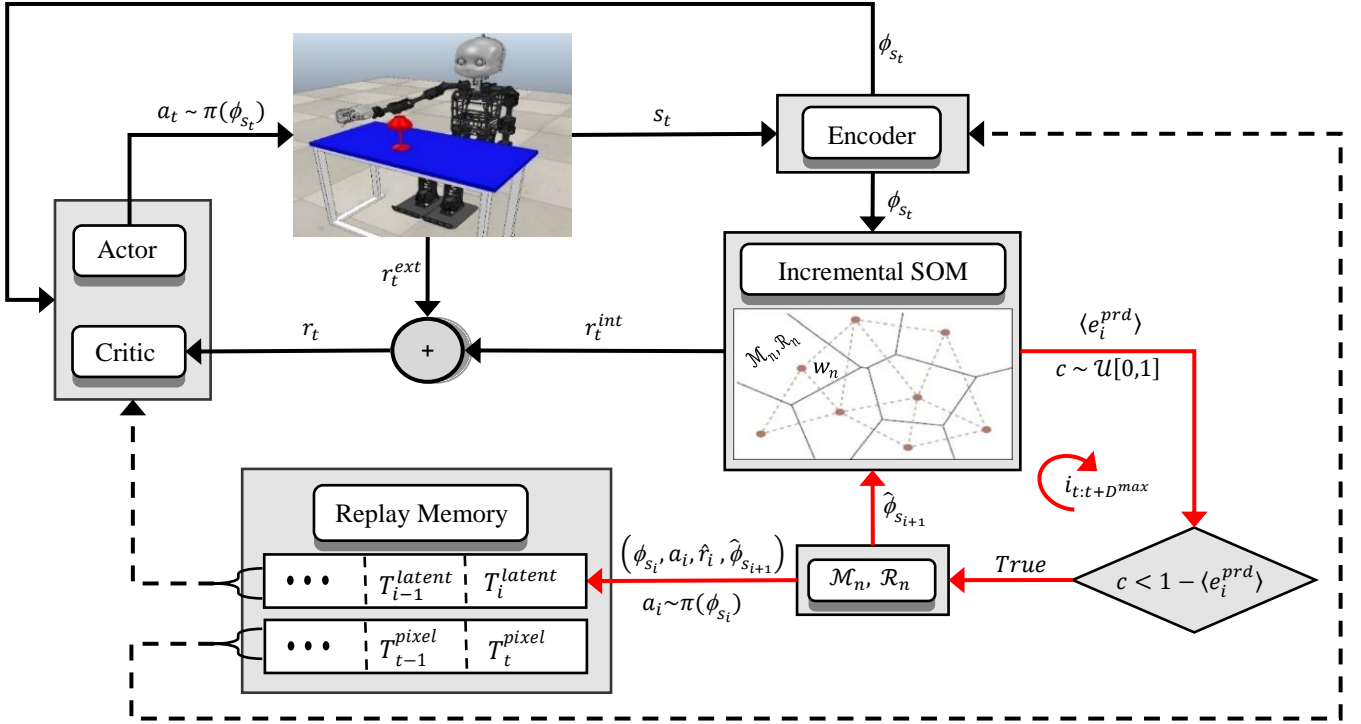

Fig. 2. Overview of our system. Solid arrows indicate information flow. Dashed arrows indicate neural network training. At each timestep *t*, the robot observes a new state $s_t$ which is then transformed into latent space with an encoder jointly trained to minimize the combined reconstruction and value prediction loss, as shown in Fig. 1. The latent encoding $\phi_{s_t}$ is used to update the incremental SOM. The best-matching node *n* identified based on the distance between its weight vector $w_n$ and $\phi_{s_t}$ (or the newly created node if the distance is greater than $e_{max}$) determines the local dynamics model $\mathcal{M}_n$ and reward function $\mathcal{R}_n$ networks associated with the region of the latent space covered by *n*. The learning parameters of $\mathcal{M}_n$ and $\mathcal{R}_n$ are then updated based on their respective predictions and the observed state transition. The combined prediction error $e_t^{prd}$ is computed and used to update the average $\langle e_t^{prd} \rangle$. An intrinsic reward $r_t^{int}$ is then derived and combined with the extrinsic reward $r_t^{ext}$ before it is fed to the critic. An on-policy imagined transition, including the predicted next encoding $\hat{\phi}_{s_{t+1}}$ and reward $\hat{r}_t$, is generated with probability inversely proportional to $\langle e_t^{prd} \rangle$. This imagined transition $T_t^{latent} = (\phi_{s_t}, a_t, \hat{r}_t, \hat{\phi}_{s_{t+1}})$ is added to the latent-space buffer of the replay memory. Again, the imagined encoding $\hat{\phi}_{s_{t+1}}$ is used to identify the next best-matching node whose dynamics and reward networks are used to generate the next imagined transition based on the respective $\langle e_t^{prd} \rangle$. This imagination process is repeated, adaptively controlled by the probability of generating imagined transitions, up to a maximum imagination depth $D^{max}$, as shown with red arrows. This is followed by updating the encoder network on a minibatch of transitions from the pixel-space buffer and the actor and critic networks on a minibatch of transitions from the latent-space buffer. Finally, the robot takes a new action sampled from the learned policy with a mean at the actor's output and a new learning cycle starts with a new observed state.

where $e_i^{prd}$ is the *i*th prediction error. We also monitor the change of average prediction error over time in each region:

$$LP_t = \left| \langle e_t^{prd} \rangle - \langle e_{t-\mathcal{W}}^{prd} \rangle \right| \quad (7)$$

where $\mathcal{W}$ is a time window. This change represents the learning progress (LP) the robot has made or expects to make.

When action $a_t$ is taken at state $s_t$, the resulting $e_t^{prd}$ associated with the best-matching node *n* of ITM (w.r.t $\phi_{s_t}$) is measured and the corresponding $\langle e_t^{prd} \rangle$ and $LP_t$ are updated. The updated $LP_t$ is then combined with the perception error $e_t^{per} = \left\| \phi_{s_{t+1}} - w_m \right\|_2^2$, where *m* is the nearest node to $\phi_{s_{t+1}}$, to produce an intrinsic reward signal:

$$r_t^{int} = LP_t + e_t^{per} \quad (8)$$

which encourages actions that maximize learning progress and lead to perceptually novel states. This is achieved by using the combined extrinsic and intrinsic reward to update the critic.

The locally trained $\mathcal{M}$ and $\mathcal{R}$ provide informative predictions with accuracy estimated by the locally stored average prediction error that can be taken into consideration when producing imagined rollouts, as explained next.

### B. Learning-adaptive (LA-) imagination

In our work, we perform imagination in latent space. To facilitate this, we split the replay memory into pixel-space and latent-space replay buffers, $B_{pixel}$ and $B_{latent}$ respectively. $B_{pixel}$ contains transitions $T_i^{pixel}$ of the form $(s_i, a_i, r_i, s_{i+1})$, while $B_{latent}$ contains transitions $T_i^{latent}$ of the form $(\phi_{s_i}, a_i, r_i, \phi_{s_{i+1}})$.

When the best-matching node *n* at timestep *t* is identified, we generate an on-policy imagined transition with probability proportional to the current accuracy of $\mathcal{M}_n$ and $\mathcal{R}_n$. This is done by first scaling the average prediction error (Eq. (6)), which is an unbiased estimate of how unreliable the recent local predictions are, to [0,1]. A number *c* is then drawn uniformly from [0,1] and an imagined transition is generated if $c < 1 - \langle e_t^{prd} \rangle$ is satisfied. The generated latent state transition $T_t^{latent} = (\phi_{s_t}, a_t, \hat{r}_t, \hat{\phi}_{s_{t+1}})$, where $\hat{r}_t = \mathcal{R}_n(\phi_{s_t}, a_t | \theta^{\mathcal{R}})$, $\hat{\phi}_{s_{t+1}} = \mathcal{M}_n(\phi_{s_t}, a_t | \theta^{\mathcal{M}})$, and $a_t \sim \pi(\phi_{s_t})$, is added to $B_{latent}$. The imagined $\hat{\phi}_{s_{t+1}}$ is used to identify the next best-matching node and the imagination process is repeated. The generation of imagined transitions fully adapts to the changes in learning local $\mathcal{M}$ and $\mathcal{R}$ networks. Similarly, the length of the imagined rollout is adaptively determined by the average prediction error in the traversed latent-space regions and is bounded by a maximum imagination depth $D^{max}$ to limit the computational time. The imagination process is detailed in Algorithm 1.

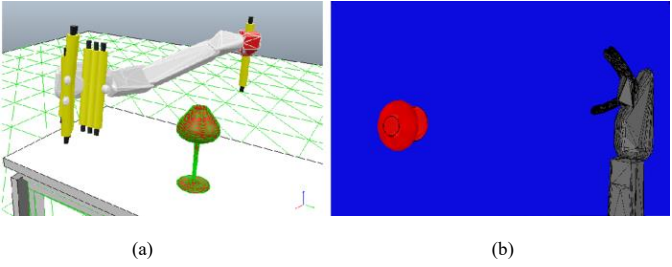

Fig. 3. (a) Motor output and (b) sensory input for the grasp learning task. The axes of rotation of the controlled joints are depicted as yellow cylinders in (a).

**Algorithm 1** LA-Imagination ($\phi_{s_t}$, $n$)

1: **Input:** Max. imagination depth $D^{max}$.
2: Scale $\langle e_t^{prd} \rangle$ of $n$ to [0,1] by its maximum over a window $\mathcal{W}$
3: $i \leftarrow 0$, $\phi_{s_i} \leftarrow \phi_{s_t}$, $\langle e_i^{prd} \rangle \leftarrow \langle e_t^{prd} \rangle$
4: Generate random number $c \sim \mathcal{U}[0,1]$
5: **while** ($c < 1 - \langle e_i^{prd} \rangle$) and ($i \leq D^{max}$) **do**
6:    Generate imagined transition $T_i^{latent} = \left( \phi_{s_i}, a_i, \hat{r}_i, \hat{\phi}_{s_{i+1}} \right)$ using $\mathcal{M}_n$ and $\mathcal{R}_n$, where $a_i \sim \pi(\phi_{s_i})$
7:    Store $T_i^{latent}$ in $B_{latent}$
8:    $\phi_{s_i} \leftarrow \hat{\phi}_{s_{i+1}}$
9:    Find best-matching node $n$ to $\phi_{s_i}$
10:   Scale $\langle e_i^{prd} \rangle$ of $n$ to [0,1] and generate $c \sim \mathcal{U}[0,1]$
11:   $i \leftarrow i + 1$
12: **end while**

Both the RL controller and the imagination process in our learning system are mutually improving, since the former is motivated to collect experiences leading to data that improves future predictions through using Deep ICAC and the latter augments the available training experiences with imagined ones in an adaptive manner to improve the sample efficiency of the former. Fig. 2 shows an overview of our learning system. We summarize the overall procedure in algorithm 2.

## V. EXPERIMENTAL RESULTS

Deep ICAC was previously found to be more stable and sample-efficient than CACLA and DDPG on visuomotor tasks [12]. Here we show the effect of incorporating learning-adaptive imagination on the sample efficiency. We evaluate our approach on learning vision-based robotic grasping.

**Parameters.** We use the learning architecture shown in Fig. 1 for approximating the policy and value functions. All convolutional layers are zero-padded, have stride 1, and use ReLU activations. All dense layers use ReLU activations except for the actor's and critic's output layers that use a tanh and a linear activation respectively. The target network's update rate $\tau$ is $10^{-3}$. The loss weighting constants $\lambda_{critic}$ and $\lambda_{rec}$ are set to 1 and 0.1 respectively. The functions $\mathcal{M}$ and $\mathcal{R}$ of each region in the latent space are jointly modeled by a feedforward neural network with three dense layers: one hidden layer of 20 tanh units and two output layers for predicting the next latent encoding and immediate reward with 16 and 1 linear units respectively. The discount factor $\gamma$ is set to 0.99. The time windows $\sigma$ and $\mathcal{W}$ are set to 40 and 20 respectively. We scale the intrinsic reward to the interval $[-1,1]$. The node creation threshold $e_{max}$ which controls the growth of the ITM map and the maximum depth of imagination $D^{max}$ are set to 6.0 and 7 respectively. We train the networks with proportional Prioritized Experience Replay (PER) [3] using Adam optimizer and learning rate $10^{-3}$ for the critic and $\mathcal{M}$ and $\mathcal{R}$ functions and $10^{-4}$ for the actor. We use two replay buffers $B_{pixel}$ of size 60K and $B_{latent}$ of size 200K, consuming 40% less memory space than the replay buffer of the Deep ICAC baseline which has a size of 100K, and a minibatch size of 64 sampled by PER. The PER parameters $\alpha$ and $\beta_0$ are set to 0.6 and 0.4 respectively. We perform 15 optimization steps on the actor and critic networks and 10 steps on $\mathcal{M}$ and $\mathcal{R}$ per timestep. We use a stochastic Gaussian policy with a mean at the actor's output and a standard deviation of 0.35 radians. Actions are capped at 20 units before being sent to the environment. The parameters of Deep ICAC are taken from our previous work [12] to ensure comparability of results. All other parameters were determined empirically through preliminary experiments.

**Algorithm 2** Deep ICAC + LA-Imagination

1: **Input:** Node creation threshold $e_{max}$, target network's update rate $\tau$, episode length $T$, no. of episodes $E$.
2: Initialize learning parameters $\{\omega, \tilde{\omega}, \theta^V, \theta^{AC}, \omega', \theta^{V'}\}$
3: Initialize incremental SOM
4: Initialize replay buffers $B_{pixel}$ and $B_{latent}$
5: **for** $e = 1$ to $E$ **do**
6:   Sample initial state $s_1$
7:   **for** $t = 1$ to $T$ **do**
8:     Compute latent state encoding $\phi_{s_t} = f(s_t \mid \omega)$
9:     Update incremental SOM
10:    Identify best-matching (or newly created) node $n$
11:    Select action $a_t \sim \pi(\phi_{s_t})$, following Eq. (3)
12:    Execute $a_t$ and observe $r_t^{ext}$ and $s_{t+1}$
13:    Compute intrinsic reward $r_t^{int}$ using Eq. (8)
14:    Compute total reward $r_t = r_t^{ext} + r_t^{int}$
15:    Update the $\mathcal{M}_n$ and $\mathcal{R}_n$ networks associated with $n$ using the transition ($\phi_{s_t}, a_t, r_t, \phi_{s_{t+1}}$)
16:    Store ($s_t, a_t, r_t, s_{t+1}$) in $B_{pixel}$ and ($\phi_{s_t}, a_t, r_t, \phi_{s_{t+1}}$) in $B_{latent}$
17:    Call LA-Imagination ($\phi_{s_t}$, $n$)
18:    Update $\{\omega, \tilde{\omega}\}$ on minibatch from $B_{pixel}$ to minimize $\mathcal{L}_{combined}$ (Eq. (5))
19:    Update $\{\theta^V, \theta^{AC}\}$ on minibatch from $B_{latent}$ to minimize $\mathcal{L}_{critic}$ (Eq. (5)), taking $\phi_{s_i}$ as an input, and $\mathcal{L}_{actor}$ (Eq. (4))
20:    Update target network parameters $\theta^{V'} \leftarrow \tau \theta^V + (1-\tau)\theta^{V'}$, $\omega' \leftarrow \tau\omega + (1-\tau)\omega'$
21:   **end for**
22: **end for**

**Results.** We compare the learning performance of Deep ICAC with and without imagination on realistic robotic grasping using V-REP robot simulator [35]. We consider two imagination types: static and learning-adaptive. The former generates imagination rollout of length $D^{max}$ at each timestep regardless of prediction errors and the latter is our proposed approach. Grasp learning is a challenging control task due to the need to perform multi-contact motions and handle rigid-body collisions with a target object. The grasping experiment here is conducted on our Neuro-Inspired COmpanion (NICO) robot [36]. NICO is a developmental humanoid built for research on neurorobotics and multimodal interaction. Fig. 2 (top-left) shows the V-REP simulated NICO in a sitting position facing a table on top of which a red glass is placed and used as a target object for grasping.

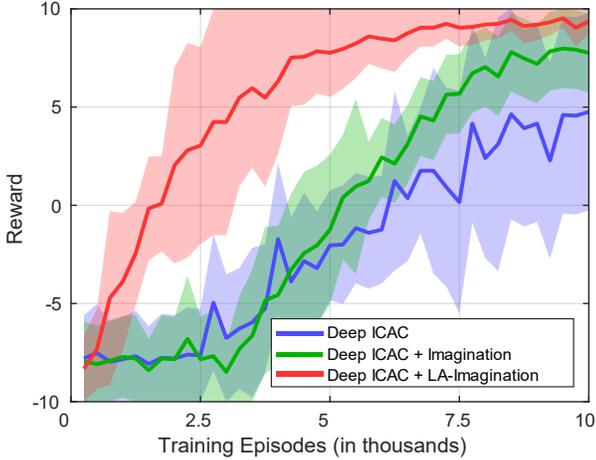

Fig. 4. Learning curves of Deep ICAC without imagination, with static imagination, and with the proposed learning-adaptive imagination on robotic grasp learning from pixels. The curves are smoothed by averaging over a moving window of 250 episodes. Shaded regions correspond to one standard deviation.

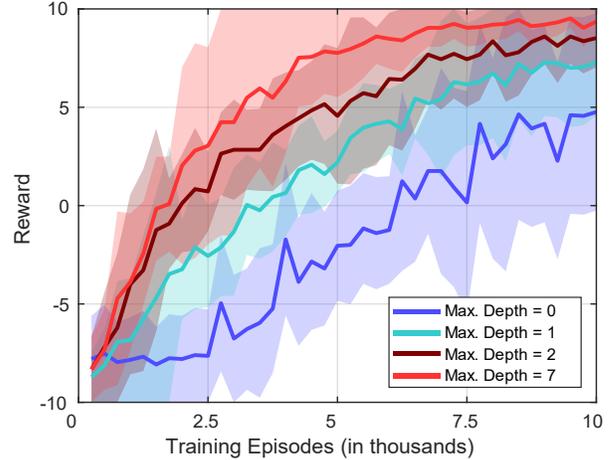

Fig. 5. Learning curves of Deep ICAC + LA-Imagination for different values of maximum imagination depth.

To avoid self-collisions while still providing a large work space for learning grasping skills, we consider a control policy involving the shoulder joint of the right arm and the finger joints of the right hand, as shown in Fig. 3(a). NICO's arm has a total of 6-Degree of Freedom (DoF) of which we control one in the shoulder with an angular range of movement of $\pm 100$ degrees. NICO's hand is 11-DoF multi-fingered with 2 index fingers and a thumb, all having an angular range of movement of $\pm 160$ degrees. NCO learns to control 2 DoFs: one for the right shoulder and one for the right hand (open/close). Each algorithm takes as its only input the 64×32 pixel RGB image from a vision sensor whose output is shown in Fig. 3(b). The reward function used is as follows:

$$r_t^{ext} = \begin{cases} +10 & if\ successful \\ -10 & if\ object\ is\ toppled \\ 0 & otherwise \end{cases}$$

To verify whether an attempted grasp was successful, the hand is closed and the shoulder joint is rotated 20 degrees in the opposite direction to that of its newly attained position. The distance between the target object's and the hand's centers is then checked if it is below a threshold of 0.04 m. If yes, the last joint position update is deemed successful. Otherwise, the hand is opened and the shoulder joint is brought back to its last position and the learning continues.

We run the algorithms on a single Nvidia GTX 1050 Ti GPU for 10K episodes and 50 steps per episode and with the target object's position randomly changing to a new graspable position at the start of each episode. The episode ends when the object is grasped, toppled, or the maximum episode length is reached. The average training time (hours) per run is 29.3±4.1 over a total of 15 runs (5 for each algorithm). Fig. 4 shows the mean episode extrinsic reward of running the algo-rithms over five random seeds. Simply using imagination irrespective of state and reward prediction accuracy resulted in poor performance, even worse than the Deep ICAC baseline, over half of the learning process, as shown in the figure. In contrast, using learning-adaptive imagination led to significantly better performance, reaching higher rewards early and converging to a near-optimal policy in less than 6K episodes. In TABLE I, we compare the learning speed and convergence (final performance) of the algorithms.

TABLE I. Learning speed (avg. reward per episode over the entire learning process) and convergence (avg. reward per episode over last 100 episodes).

|  | Deep ICAC | Deep ICAC + Imagination | Deep ICAC + LA-Imagination |
|---|---|---|---|
| Learning speed | -2.039 | -0.548 | **5.571** |
| Convergence | 5.4 | 8.3 | **9.4** |

We also evaluate the effect of using different values of maximum imagination depth $D^{max}$ on the performance. Fig. 5 shows the mean episode extrinsic reward of Deep ICAC+LA-Imagination on the visual grasping task for different maximums of imagination depth, averaged over five runs. A rollout of a single imagined step was enough to improve the performance over the baseline (no imagination). Similarly, going from a maximum of one to two imagined steps allowed faster learning in the early episodes and led to a better final policy. Seven outperformed two, reaching higher reward after just 2K episodes. Values greater than seven did not change the performance. This is because the length of the imagined rollout is often shorter than a large $D^{max}$, as it stops increasing when the model prediction error is high before reaching $D^{max}$.

## VI. CONCLUSION

This paper establishes a bridge between intrinsic motivation and imagination in robot decision-making, inspired by human mental simulation of motor behavior. Our approach performs imagination in high-level latent space, resembling human imagination operating on abstract representations, to provide additional training experiences and accelerate skill learning. Unlike previous works, our approach generates imagined experiences only when the learned dynamics and reward functions have high local prediction accuracy, thus adapting to the learned underlying dynamics. In our approach, the imagina-tion depth is adaptively determined using spatially and temporally local information provided by the average prediction error computed in different regions of the latent space over a recent time interval. We showed that integrating our approach to imagination with dynamics-based intrinsic motivation makes learning pixel-level control policies more efficient, particularly for robotic grasping in sparse reward environment. Future work will focus on evaluating the proposed approach on visually more complex environments and different grasp-learning objects. We also plan to investigate how the trained policies perform on the physical robot.

ACKNOWLEDGEMENT

This work was supported by the DAAD German Academic Exchange Service (Funding Programme No. 57214224) with partial support from the German Research Foundation DFG under project CML (TRR 169).